\documentclass[letterpaper, 10 pt, conference]{ieeeconf}  

\IEEEoverridecommandlockouts                              

\usepackage{graphicx}
\usepackage{amsfonts}
\usepackage{comment}
\usepackage{url}
\usepackage[ruled,vlined,linesnumbered]{algorithm2e}
\usepackage[acronym]{glossaries}
\usepackage{booktabs}
\usepackage{multirow}
\usepackage{subcaption}
\usepackage{eso-pic}
\usepackage[dvipsnames]{xcolor}
\usepackage[capitalise]{cleveref}

\DeclareMathAlphabet\mathbfcal{OMS}{cmsy}{b}{n}
\newacronym{dof}{DoF}{Degrees of Freedom}
\newacronym{adf}{ADF}{Anchor Description Format}
\newacronym{ae}{AE}{Auto-Encoder}
\newacronym{pca}{PCA}{Principal Component Analysis}
\newacronym{cvae}{CVAE}{Conditional Variational Auto-Encoder}
\newacronym{icp}{ICP}{Iterative Closest Point}
\newacronym{rl}{RL}{Reinforcement Learning}

\title{\LARGE \bf
PCHands: PCA-based Hand Pose Synergy\\Representation on Manipulators with \textit{N}-DoF
}

\author{En Yen Puang$^{1,2}$, Federico Ceola$^{1}$, Giulia Pasquale$^{1}$ and Lorenzo Natale$^{1}$
\thanks{$^{1}$En Yen Puang, Federico Ceola, Giulia Pasquale, and Lorenzo Natale are with Humanoid Sensing and Perception (HSP), Istituto Italiano di Tecnologia (IIT), Genoa, Italy (email: {\tt\footnotesize name.surname@iit.it}).}
\thanks{$^{2}$En Yen Puang is also with Dipartimento di Informatica, Bioingegneria, Robotica e Ingegneria dei Sistemi, University of Genoa, Genoa, Italy.}
}

\newcommand{\firstpagecopyright}{
    \AddToShipoutPictureFG*{
        \AtPageUpperLeft{
        \hspace*{\dimexpr1in+\oddsidemargin\relax}
        \raisebox{-3.5\baselineskip}[0pt][0pt]{
            \begin{minipage}{\textwidth}
            \centering\footnotesize
            \textit{\textcopyright~2025 IEEE. Personal use of this material is permitted.
            Permission from IEEE must be obtained for all other uses, in any current or future media,
            including reprinting/republishing this material for advertising or promotional purposes,
            creating new collective works, for resale or redistribution to servers or lists, or reuse of
            any copyrighted component of this work in other works.} \\
            Preprint version (Oct.\ 2025). This work has been accepted for publication in 2025 IEEE-RAS 24th International Conference on Humanoid Robots.
            \end{minipage}
        }
        }
    }
}

\begin{document}
\firstpagecopyright
\maketitle
\thispagestyle{empty}
\pagestyle{empty}

\begin{abstract}
We consider the problem of learning a common representation for dexterous manipulation across manipulators of different morphologies.
To this end, we propose PCHands, a novel approach for extracting hand postural synergies from a large set of manipulators. 
We define a simplified and unified description format based on anchor positions for manipulators ranging from 2-finger grippers to 5-finger anthropomorphic hands. 
This enables learning a variable-length latent representation of the manipulator configuration and the alignment of the end-effector frame of all manipulators. 
We show that it is possible to extract principal components from this latent representation that is universal across manipulators of different structures and degrees of freedom. 
To evaluate PCHands, we use this compact representation to encode observation and action spaces of control policies for dexterous manipulation tasks learned with \acrfull{rl}. 
In terms of learning efficiency and consistency, the proposed representation outperforms a baseline that learns the same tasks in joint space.
We additionally show that PCHands performs robustly in \acrshort{rl} from demonstration, when demonstrations are provided from a different manipulator. 
We further support our results with real-world experiments that involve a 2-finger gripper and a 4-finger anthropomorphic hand.
Code and additional material are available at \url{https://hsp-iit.github.io/PCHands/}.
\end{abstract}

\section{Introduction}

The wide availability of large heterogeneous datasets has been key to training models with impressive capabilities in \textit{Natural Language Processing} (NLP)~\cite{achiam2023gpt} and \textit{Computer Vision} (CV)~\cite{radford2021learning, kirillov2023segany}. The lack of such datasets in robotics is one of the factors hampering the development of a large generalist model for robotic tasks. There is a huge effort in the robotics community to collect large-scale datasets~\cite{vuong2023open, khazatsky2024droid} and train generalist models~\cite{octo_2023, kim24openvla} with these data. However, these datasets usually share two limitations: their size (0.16M tasks~\cite{vuong2023open}) is orders of magnitude smaller than those used for NLP (1.5-4.5B)~\cite{lehmann2015dbpedia, muhleisen2012web} and CV (5-18M)~\cite{weyand2020google, wu2019tencent} tasks, and they mostly target tasks carried out with two-finger grippers.

When considering dexterous manipulation tasks such as in-hand re-orientation~\cite{chen2022system} and tool-use~\cite{Rajeswaran2018dapg}, the availability of data is even more limited, and the difference in manipulator structures hinders the deployment of a large dataset for these tasks. In contrast, there is a wealth of human datasets performing everyday tasks that require a high level of dexterity, which could potentially be used to teach a robot to solve such tasks. 

\begin{figure}[t]
  \vspace{2mm}
  \begin{center}
    \includegraphics[width=0.9\linewidth]{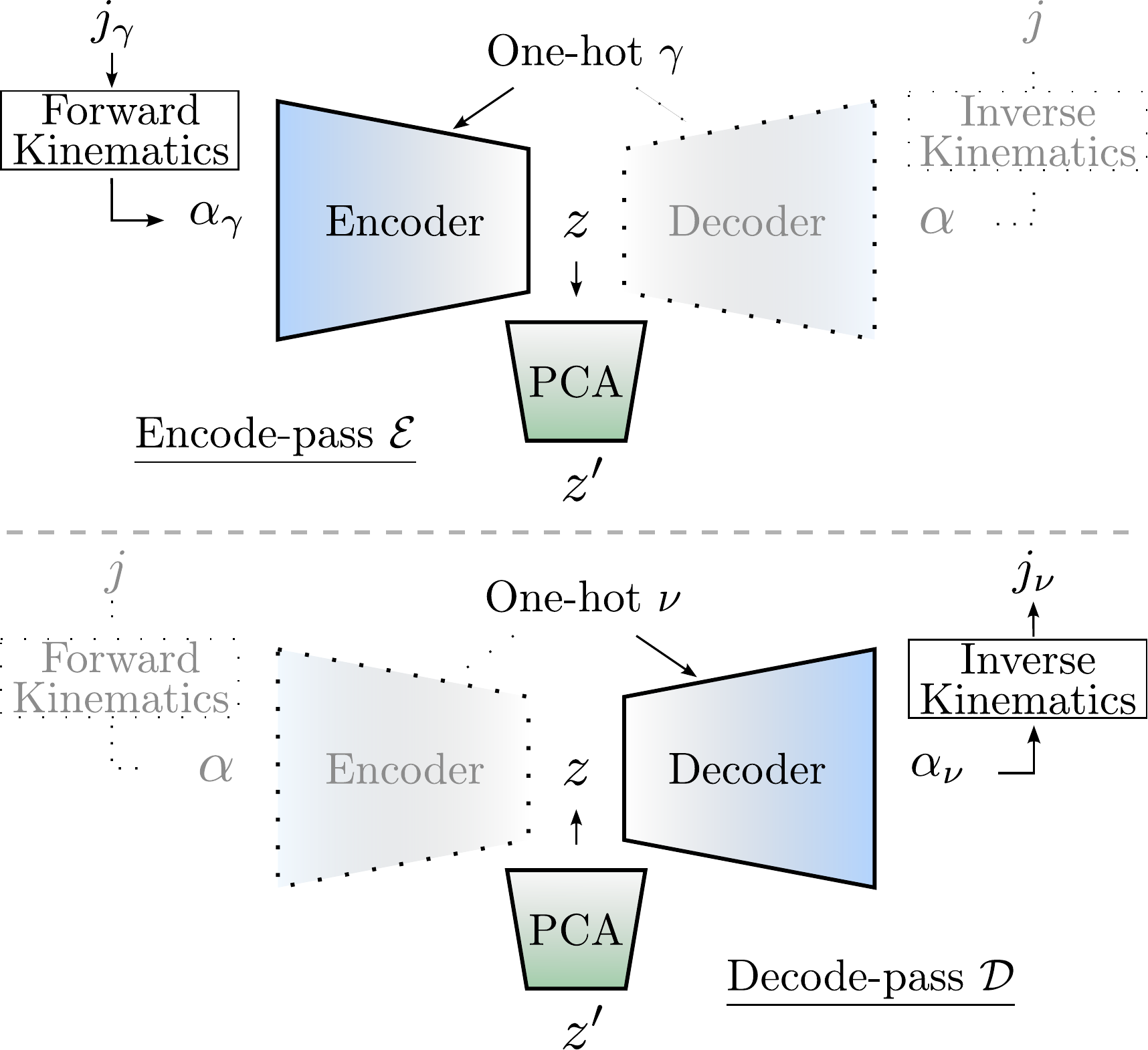}
  \end{center}
  \vspace{-2mm}
  \caption{The proposed architecture consists of a \acrshort{cvae} which encodes anchors \textit{$\alpha$} into (and decodes them from) a latent synergy space \textit{z} conditioned on the one-hot manipulator identifier, and a linear \acrshort{pca} to extract the most significant \acrshort{dof} $z'$ in representing poses under \acrshort{adf}. The architecture can be used to retarget manipulator poses ($j_\gamma$ to $z$ or $z'$ with encode-pass, followed by $z$ or $z'$ to $j_\nu$ with decode-pass) or to 
  directly control manipulator $\nu$ from the common variable-length representation ($z'$ to $j_\nu$) with decode-pass.}
  \label{fig:model}
  \vspace{-3mm}
\end{figure}

To overcome these limitations, we propose PCHands, a novel \acrshort{pca}-based method for extracting a common representation of hand postural synergies spanning human, dexterous robotic hands, and two-finger grippers.
PCHands learns a variable-length postural synergy representation across a large set of very different manipulators by means of three components: the proposed \acrfull{adf}, \acrfull{cvae}~\cite{sohn2015learning}, and \acrfull{pca}~\cite{abdi2010principal}.
In \acrshort{adf}, we first position a predefined set of anchors (also known as keypoints) on each manipulator. We then collect a dataset of anchor positions for all manipulators at random joint configurations.
Next, we devise a procedure that iteratively (i) trains a \acrshort{cvae} on the anchors dataset to extract a common latent representation of the postural synergies, and (ii) aligns the end-effector of each manipulator with \acrfull{icp}. Moreover, we propose to further process the latent representation with \acrshort{pca} to extract a variable-length latent representation. This provides flexibility in dimensionality selection and allows for accommodating different levels of complexity in both manipulators and downstream tasks.

We show that PCHands finds unified and coherent postural synergies of a wide range of manipulators with diverse morphologies and \acrfull{dof}. Concretely, we show that the first principal component of the latent representation corresponds to a universal hand-opening motion across 17 manipulators. 
We use PCHands for online retargeting of task demonstrations collected via teleoperation in simulation, from a human hand to robotic manipulators. Subsequently, we leverage the latent synergy representation extracted with PCHands to train \acrshort{rl} policies on different tasks with several manipulators having different \acrshort{dof}.
We show that PCHands improves the training of \acrshort{rl} by using the common representation in both the observation and action spaces. Specifically, we benchmark PCHands and a state-of-the-art baseline on 5 dexterous manipulation tasks. Overall, PCHands achieves better learning efficiency and consistency over the baseline that learns policies in joint space.

Finally, we evaluate the coherence and robustness of the PCHands unified representation by using demonstrations collected with different manipulators to train manipulation tasks with \acrshort{rl}. We show that PCHands is comparable and consistent across different combinations of source demonstrations and target manipulators. To show the applicability in real-world scenarios, we also roll-out policies trained in simulation on two manipulators on a 7-\acrshort{dof} robotic arm. Although having a subtle performance decline, PCHands still shows promising applicability in real-world scenarios. 

In summary, the contributions of this work are:
\begin{itemize}
    \item We propose PCHands as an architecture to learn a common, variable-length representation of hand postural synergies for a wide range of different manipulators. 
    \item We propose \acrshort{adf} that enables the extraction of the latent postural synergy representation with \acrshort{cvae} and \acrshort{pca}, and alignment of end-effector frames with \acrshort{icp}. 
    \item We demonstrate the advantages of PCHands in learning dexterous manipulation tasks with \acrshort{rl}. PCHands shows better consistency and faster convergence than the state-of-the-art and offers flexibility in data collection. 
\end{itemize}

\section{Related Work}

\subsection{Hand Postural Synergies by Dimensionality Reduction}
Dimensionality reduction is a widely used approach to extract postural synergies of robotic manipulators. It compresses raw joint values $j$ into a lower-dimensional space $z$ with $\mathrm{dim}(z) \leq \mathrm{dim}(j)$. Dimensionality reduction is usually composed of a forward mapping $q_\phi: j \mapsto z$ that \textit{encodes} joint values into the compressed latent representation and a backward mapping $p_\theta: z \mapsto j$ that \textit{decodes} the compressed values back to the joint representation. Two different approaches to reduce the dimensionality of the latent space are \acrshort{pca}~\cite{pearson1901liii} and GPLVM~\cite{lawrence2003gaussian}. \acrshort{pca} is a linear method that decomposes data into principal components ranked according to the variance, such that discarding the least significant principal components reduces the dimensionality of the latent space with minor information loss. \acrshort{pca} has been used in~\cite{ciocarlie2007dimensionality, bernardino2013precision} to find a low-dimensional basis for robotic grasp postures and improve optimization of stable grasp search for complex manipulators. Gaussian Process Latent Variable Model (GPLVM)~\cite{lawrence2003gaussian} instead is a non-linear probabilistic model for \acrshort{pca}, and is used in~\cite{xu2016comparative} to improve postural reconstruction error.

Deep models such as \acrshort{ae}s~\cite{kramer1991nonlinear} and \acrshort{cvae}s~\cite{sohn2015learning} have also recently been adopted to learn postural synergies of manipulators~\cite{starke2018synergy, dimou2023robotic}. The objective of a \acrshort{cvae} is to minimize the reconstruction error with an \acrshort{ae}, while a Kullback-Leibler divergence term regularizes the latent space to a Gaussian prior. Both the encoder and the decoder of the \acrshort{cvae} are conditioned on the task variable $c$, and the objective function to maximize is defined as:
\begin{multline} 
    \label{equ:cvae}
    \mathcal{L}_{\theta, \phi}(x,c) = \mathbb{E}_{z\sim q_\phi(z|x,c)}\big [\text{log}\, p_\theta (x|z,c) \big ] \\
    - \lambda D_{KL}\big( q_\phi(z|x,c) \| p_\theta(z) \big)
\end{multline}

The approach proposed in~\cite{starke2018synergy} uses an AE to extract grasping synergies from a human grasping dataset. In~\cite{dimou2023robotic}, instead, a \acrshort{cvae} is applied to emphasize the smoothness of the latent space when interpolating between different latent configurations, while conditioning the synergies with task-related variables. Although having better performance than \acrshort{pca} variants, these deep learning-based methods do not naturally offer variable-length flexibility in the latent representation. To this end, we propose a combination of non-linear \acrshort{cvae} and linear \acrshort{pca} to retain the representational power of a deep model while staying flexible in the size of the latent representation. 

\subsection{Hand Pose Representation with Anchors}
Anchors are introduced in~\cite{yang2021cpf} to simplify the representation of hand meshes~\cite{romero2017embodied}. Anchors are a set of points that are positioned deliberately on a hand and used to distinctly represent a subdivision of hand regions, e.g., distal, proximal, and metacarpal. The use of a subdivided hand representation is supported by the analysis of human manipulation datasets~\cite{fan2023arctic, kwon2021h2o}, which show the differences in contact frequency across hand regions. 

Grasp pose optimization in~\cite{ciocarlie2007dimensionality, yang2021cpf} uses anchors to simplify the representation of hand-object contacts. Similarly, hand pose retargeting methods proposed in~\cite{antotsiou2018task, qin2022one, qin2022dexmv, qin2023anyteleop} model the problem using forward and inverse kinematics passes of hand key-points, and solve it with optimization-based approaches. Particle Swarm Optimization~\cite{kennedy1995particle}, Quadratic Programming~\cite{nlopt2007} algorithms, and differentiable kinematics~\cite{pinocchio2015web} are usually employed in the optimization with the following objective:
\begin{equation}
    \label{equ:optim}
    \min_{j_t}\big \| \alpha^* - f(j_t) \big \| + \lambda \big \| j_{t-1} - j_t \big \|,
\end{equation}
where $\alpha^*$ is the position of anchors on the source manipulator, and $f(\cdot)$ is the forward kinematics function that projects joint values $j_t$ to anchors on the target manipulator. $\lambda$ is used to weigh temporal consistency for a minimal joint jerk.

In this work, we adopt an anchor-based hand pose representation and stretch its application by using the proposed \acrshort{adf} not only on anthropomorphic hands but also on manipulators with different numbers of fingers and structures. Notably, PCHands does not require optimization post-processing for pose retargeting and relies only on forward-inverse kinematics, encode-decode passes of a \acrshort{cvae}, and forward-inverse transforms of linear \acrshort{pca}. These components are efficient and can be run online.

\begin{figure}[t]
  \vspace{2mm}
  \begin{center}
    \includegraphics[width=0.80\linewidth]{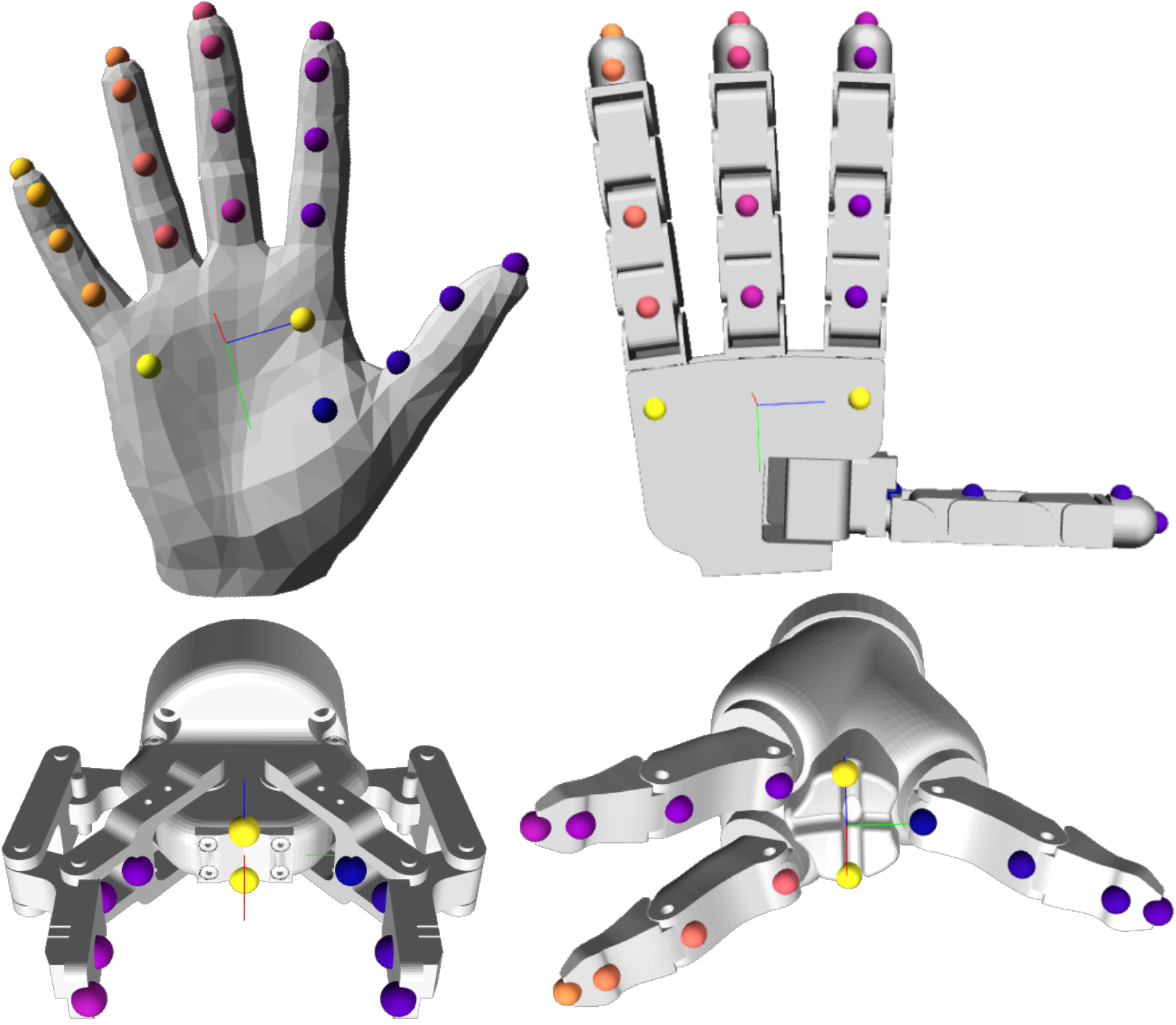}
  \end{center}
  \vspace{-2mm}
  \caption{Position of the set of 22 anchors on human and anthropomorphic hands, and 2 \& 3-finger grippers. Each color-coded anchor under \acrshort{adf} carries symbolic meaning about the region it represents consistently across manipulators.}
\label{fig:anchors}
\vspace{-3mm}
\end{figure}

\section{Methodology}

PCHands first standardizes the geometry of manipulators using a set of predefined anchors via \acrshort{adf} (Sec.~\ref{sec:adf}). Then, a two-stage dimensionality reduction model (Sec.~\ref{sec:synergy}) is used to extract a variable-length low-dimensional latent representation of manipulator poses. Finally, an iterative learning procedure (Sec.~\ref{sec:refine}) refines end-effector frames and ensures a morphologically coherent representation across a diverse set of manipulators.

\subsection{\acrlong{adf}}
\label{sec:adf}

To create a unified representation of the configuration of different manipulators, the proposed \acrshort{adf} extends the use of anchors beyond representing anthropomorphic hands, to parallel and multi-finger grippers. 

\noindent{\textbf{Anchors Placement}} 
We define an ordered set of 22 anchors $\alpha = \{ x_i \ | \ x_i \in \mathbb{R}^{3} \}_{i=1}^{22}$, where each $x_i$ denotes a 3D point placed on a predefined functional part of a manipulator. These anchors are manually positioned on each manipulator, as shown in~\cref{fig:anchors}. In 5-finger anthropomorphic hands, 4 anchors are placed on each finger (at the proximal, intermediate, distal, and tip phalanges), and 2 on the palm. For 2-finger grippers, 4 thumb anchors are assigned to the left jaw, and the remaining 16 finger anchors are merged on the right jaw. This ``anchor-merging" approach generalizes to manipulators with fewer than five fingers. Palm anchors are placed centrally at the gripper base.

\noindent{\textbf{Preliminary End-effector Frame Placement}} 
We define a preliminary end-effector frame between the two palm anchors, with the x-axis pointing outward from the palm and the y-axis toward the wrist (hands) or thumb jaw (grippers). All anchor positions are expressed relative to this frame. However, morphological differences cause inconsistency in this frame across manipulators, limiting representation coherence under the \acrshort{adf}. To address this issue, we refine the placement of the end-effector frame as detailed in Sec.~\ref{sec:refine}.

\begin{figure}[t]
  \vspace{2mm}
  \begin{center}
    \includegraphics[width=1\linewidth]{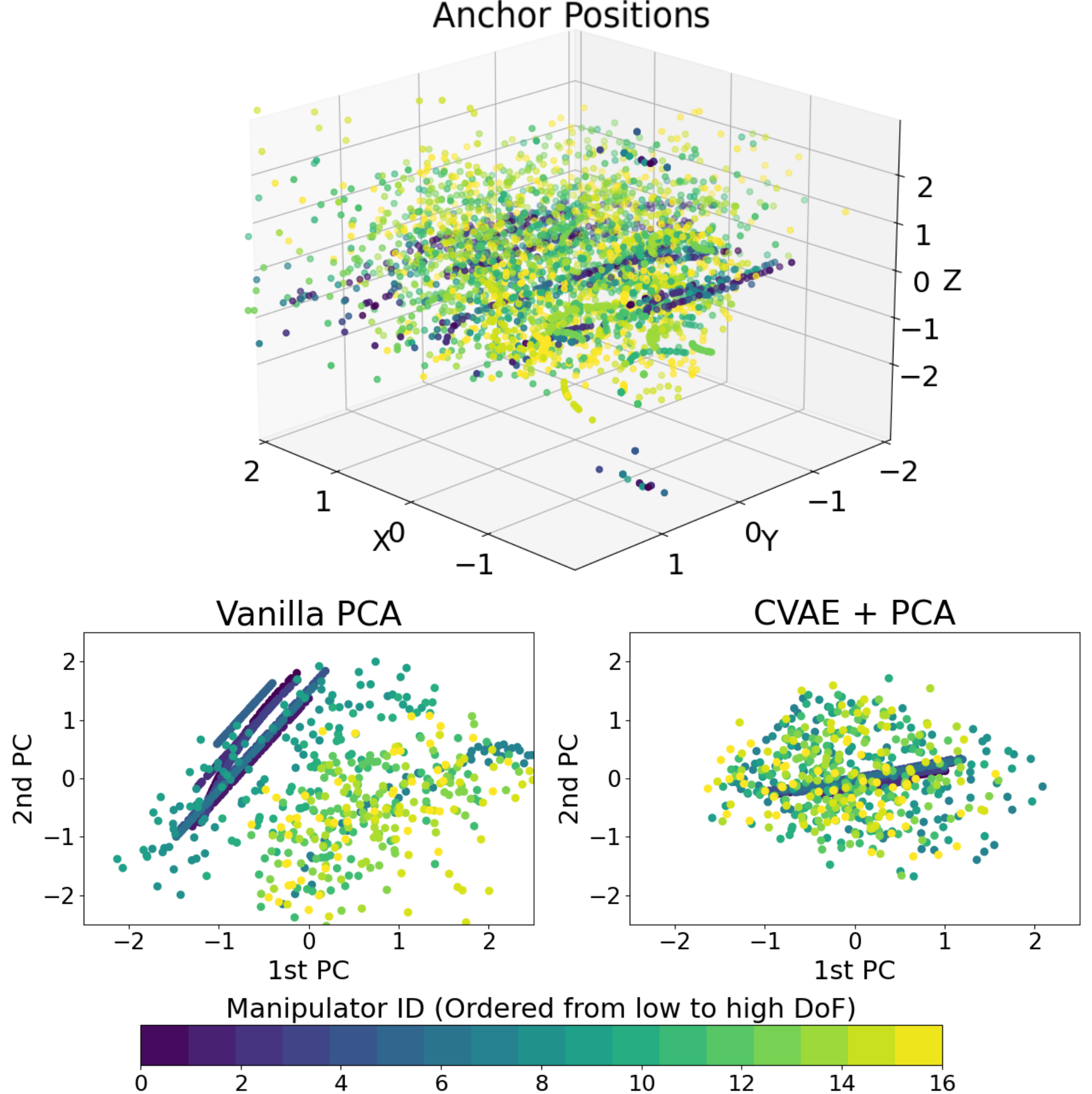}
  \end{center}
  \vspace{-2mm}
  \caption{(Top) Normalized anchor positions from manipulators with various \acrshort{dof}. Two-finger grippers (Dark Blue) make mostly planar motions on the X-Y plane, while the others move in all three dimensions. (Bottom) The first and second principal components (PC) of the corresponding manipulator poses from vanilla \acrshort{pca} and our method, \acrshort{cvae}+\acrshort{pca}. Vanilla \acrshort{pca} clusters manipulators according to the morphology, losing representational capacity of hand synergies. 
  }
  \label{fig:pca}
  \vspace{-3mm}
\end{figure}

\subsection{Postural Synergy Model}
\label{sec:synergy}
We extract a variable-length low-dimensional hand pose representation coherently for each manipulator by applying a two-stage dimensionality reduction. As shown in~\cref{fig:model}, our model concatenates a \acrshort{cvae} and a linear \acrshort{pca}, allowing the encoding and decoding between anchor positions and principal component coefficients.

\noindent{\textbf{CVAE}} 
We use a \acrshort{cvae} to reduce the dimensionality of the \acrshort{adf}, encoding a set of anchor positions $\alpha$ as a latent variable $z$, where $\mathrm{dim}(z) \ll 22 \times 3$. As shown in~\cite{dimou2023robotic}, this latent space smoothly represents dexterous manipulator poses. Thus, we apply a \acrshort{cvae} to reconstruct anchor positions conditioned on a one-hot vector identifying the manipulator, used in both the encoder and decoder. The model minimizes a weighted L1 loss between input anchors $x_i$ and their reconstruction $\hat{x}_i$:
\begin{equation}
    \label{equ:vae}
    \min_{\phi, \theta} \sum_{i=1}^{22} \big  | w_i(x_i - \hat{x}_i) \big |.
\end{equation}
Weights $w_i$ are heuristically set based on the usage of anchor merging in the training dataset. Rarely merged anchors (e.g., thumb-anchors) have higher weights to ensure balanced reconstruction (e.g., a 2-finger gripper has 4 thumb-anchors on the left-jaw and 16 finger-anchors on the right-jaw).

\noindent{\textbf{Training Dataset}}
We train the \acrshort{cvae} using a dataset $\mathcal{A} = \{ \alpha\ |\ \alpha \in \mathbb{R}^{22 \times 3} \}^{ m\times n}$ composed of anchors from $m$ manipulators, ranging from 2-finger grippers to 5-finger anthropomorphic hands. For each manipulator, $n$ configurations are generated by uniformly sampling joint positions within kinematics constraints (e.g., joint limits, actuation). The corresponding anchor positions computed via forward kinematics are expressed in the updated end-effector frame (which is iteratively refined as explained in Sec.~\ref{sec:refine}) and normalized to a unit Gaussian in the 3D Cartesian space. ~\cref{fig:pca} (Top) illustrates the normalized anchor positions.

\noindent{\textbf{PCA Reduction}} 
To prevent redundancy in low-\acrshort{dof} manipulators (e.g., most 2-finger grippers have only 1 \acrshort{dof}), we apply linear \acrshort{pca} to decompose and further reduce the dimensionality of the latent variable $z$ into principal component coefficients $z'$. 
While \acrshort{pca} alone can reduce the Cartesian anchor space, it has a key limitation: as shown in~\cref{fig:pca}, the first principal component often over-represents manipulator differences over finger poses. Depending on the composition of the dataset, this na\"ive approach loses part of its capacity to represent manipulator-specific information. In contrast, our method uses a \acrshort{cvae} to model inter-manipulator variation, allowing \acrshort{pca} to focus on pose variation across all components.

\noindent{\textbf{Encode Pass}} 
The \textit{encode-pass} $ \mathcal{E}: j \mapsto \alpha \mapsto z \mapsto z'$ transforms joint values $j$ into a compact principal component representation. We first compute anchor positions $\alpha$ in the end-effector frame via forward kinematics, and encode them into the latent representation $z$ via the \acrshort{cvae} encoder. We then extract the \acrshort{pca} coefficients $z'$ as a variable-length latent representation for downstream applications.

\noindent{\textbf{Decode Pass}} 
The \textit{decode-pass} $ \mathcal{D}: z' \mapsto z \mapsto \alpha \mapsto j$ instead transforms the compact principal component representation into joint values. 
First, we compute the \acrshort{cvae}'s latent vector via inverse \acrshort{pca}. We then reconstruct the anchor positions via the \acrshort{cvae} decoder, and retrieve the joint values via multi-objective inverse kinematics for all anchors. 

\noindent{\textbf{Separation of Synergies and Hardware}} 
The proposed approach decouples the synergy model from hardware layers responsible for forward and inverse kinematics, enabling hardware-agnostic behavior. Pose retargeting is facilitated via a shared latent representation and distinct hardware layers during encoding and decoding. Specifically, to retarget poses from manipulator $\gamma$ to manipulator $\nu$, their respective hardware layers are applied in the encode and decode passes (see~\cref{fig:model}), while the rest of the model remains unchanged:
\begin{equation}
    \label{equ:retarget}
    j_\nu = \mathcal{D} \big (\nu, \mathcal{E}(\gamma, j_\gamma) \big ).
\end{equation}

\subsection{Refinement of End-effector Frame}
\label{sec:refine}

\begin{algorithm}[t]
    \DontPrintSemicolon
    \SetKwInOut{Input}{Input}
    \SetKwInOut{Output}{Output}
    \Input{$M$ manipulators in \acrshort{adf}}
    \Output{$\psi$ synergy model, $\delta$ frame alignment}
    \BlankLine
    $\delta_0 \gets \mathbf{0} $\;
    \While{$i \leq \textsc{budget}$}{ 
        $\mathcal{A}_i \gets \mathrm{create\_dataset}(\delta_i)$ \;
        $\psi_i \gets \mathrm{train\_model}(\mathcal{A}_i)$ \;
        \ForEach{$M$}{
        $\delta_{i+1} \gets \mathrm{refine\_frame}(\psi_i)$ \;
        }
    }
    \Return{$\psi$, $\delta$}
    \caption{Iterative synergy learning algorithm with end-effector frame refinement.}
    \label{algo:iterative}
\end{algorithm} 

As discussed in Sec.~\ref{sec:adf}, the preliminary placement of end-effector frames must be refined individually to address morphological differences. We introduce an iterative process that alternates between refining end-effector frames and retraining the synergy model. 
Under the same end-effector pose and latent finger configuration, the algorithm ensures optimal alignment in the end-effector frame and anchor position for all manipulators.

\noindent\textbf{Iterative Learning Procedure}
As illustrated in Algo.~\ref{algo:iterative}, the proposed iterative learning procedure runs within a computation \textsc{budget} (number of iterations). It iterates between fitting the synergy model ($\mathrm{create\_dataset}(\cdot)$ and $\mathrm{train\_model}(\cdot)$) as described in Sec.~\ref{sec:synergy} and refining the end-effector frames ($\mathrm{refine\_frame}(\cdot)$) as described in Algo.~\ref{algo:refinement}. 
The iterative nature of the algorithm ensures that the synergy model $\psi$ is always trained with the dataset $\mathcal{A}$ that refers to the latest end-effector frames that have been adjusted by the new refinement $\delta_i$.

\noindent{\textbf{Anchors Alignment}} 
The refinement procedure described in Algo.~\ref{algo:refinement} requires a subset of manipulators referred to as reference manipulators $\mathrm{ref}$. They provide the reference anchors to which anchors from the target manipulator $\mathrm{tgt}$ align. To attain stable performance, the selection criteria for reference manipulators are simplicity and morphological diversity. Hence, we pick \textit{Robotiq-2f85}, \textit{Google-gripper}, \textit{Kinova-3f} and \textit{Armar-hand}. This refinement procedure is applied to all manipulators, including the ones in the $\mathrm{ref}$ subset.

\noindent First, the algorithm samples $k$ evenly spaced points on the first principal component (Lines 1-2). For each point, the \textit{decode-pass} is used to map the sampled principal component coefficient to anchors on the target (Line 3) and reference (Lines 4-5) manipulators. A set of average anchor positions is then computed among the reference manipulators (Line 6). 
The above procedure is repeated for each sampled point on the first principal component (Lines 1-6). Lastly, the direct correspondences between target and reference anchors in each of these $k$ configurations are used to compute the adjustment on the end-effector frame of the target manipulator. The optimal rigid transformation $\delta = \{ R, t\} \in$ SE(3) is computed using a single step of \acrshort{icp}~\cite{besl1992method} with objective
\begin{equation}
    \label{equ:icp}
    \min_{R,\ t}\sum_{i=1}^k \big \| \alpha_i^\mathrm{ref} - R \alpha_i^\mathrm{tgt} - t \big \|^2
\end{equation}
and solved via weighted singular value decomposition. Higher weights are given to the fingertips and thumb anchors for a more balanced alignment. 

\begin{algorithm}[t]
    \DontPrintSemicolon
    \SetKwInOut{Input}{Input}
    \SetKwInOut{Output}{Output}
    \Input{$\psi$ synergy model, $\mathrm{ref}$ and $\mathrm{tgt}$ manipulators}
    \Output{$\delta$ frame alignment}
    \BlankLine
    \For{$i \in \mathrm{linspace}(\pm2.5,\ k)$}{
        \label{line:linspace}
        $z'_i = [i, 0, \cdots, 0]$ \tcc*[r]{$1^{st}$ PC}
        $\alpha_i^\mathrm{tgt} \gets \mathcal{D}_\psi(\mathrm{tgt}, z'_i)$ \;
        \For{$j \in \mathrm{ref}$} 
        {
            $\alpha_i^{\mathrm{ref}_j} \gets \mathcal{D}_\psi(\mathrm{ref}_j, z'_i)$ \;
        }
        $\alpha_i^\mathrm{ref} \gets \mathrm{mean}_j(\alpha_i^{\mathrm{ref}_j}) $ \;
    }
    $\delta^\mathrm{tgt} \gets \mathrm{ICP}\big(\mathrm{concat}_i(\alpha_i^\mathrm{tgt}),\ \mathrm{concat}_i(\alpha_i^\mathrm{ref}) \big) $ \;
    \Return{$\delta^\mathrm{tgt}$}
    \caption{End-effector Frame Refinement}
    \label{algo:refinement}
\end{algorithm}

\section{Experiments}
\label{sec:experiments}
We include $m=17$ manipulators in PCHands, with each manipulator contributing $n=10000$ configuration samples to the dataset $\mathcal{A}$.
The encoder and decoder of our \acrshort{cvae} are implemented using a 4-layer MLP with layer normalization. 
We set the dimension of the \acrshort{cvae} latent space $\mathrm{dim}(z) = 10$ in all experiments, enabling selection of 1–10 principal components as the final \acrshort{dof}. For further details, we refer the reader to the code, model checkpoint, and collected demonstrations available on the project page.

\subsection{Qualitative Analysis of the Synergy Model}

As depicted in~\cref{fig:pca} (Bottom Right), poses from low-\acrshort{dof} manipulators tend to spread more in the first principal component than the others. This suggests that the most significant principal component found by our approach corresponds to the dimension that is universal in representing simple opening motions across manipulators with different \acrshort{dof}.

\begin{figure}[t]
  \vspace{2mm}
  \begin{center}
    \includegraphics[width=1.0\linewidth]{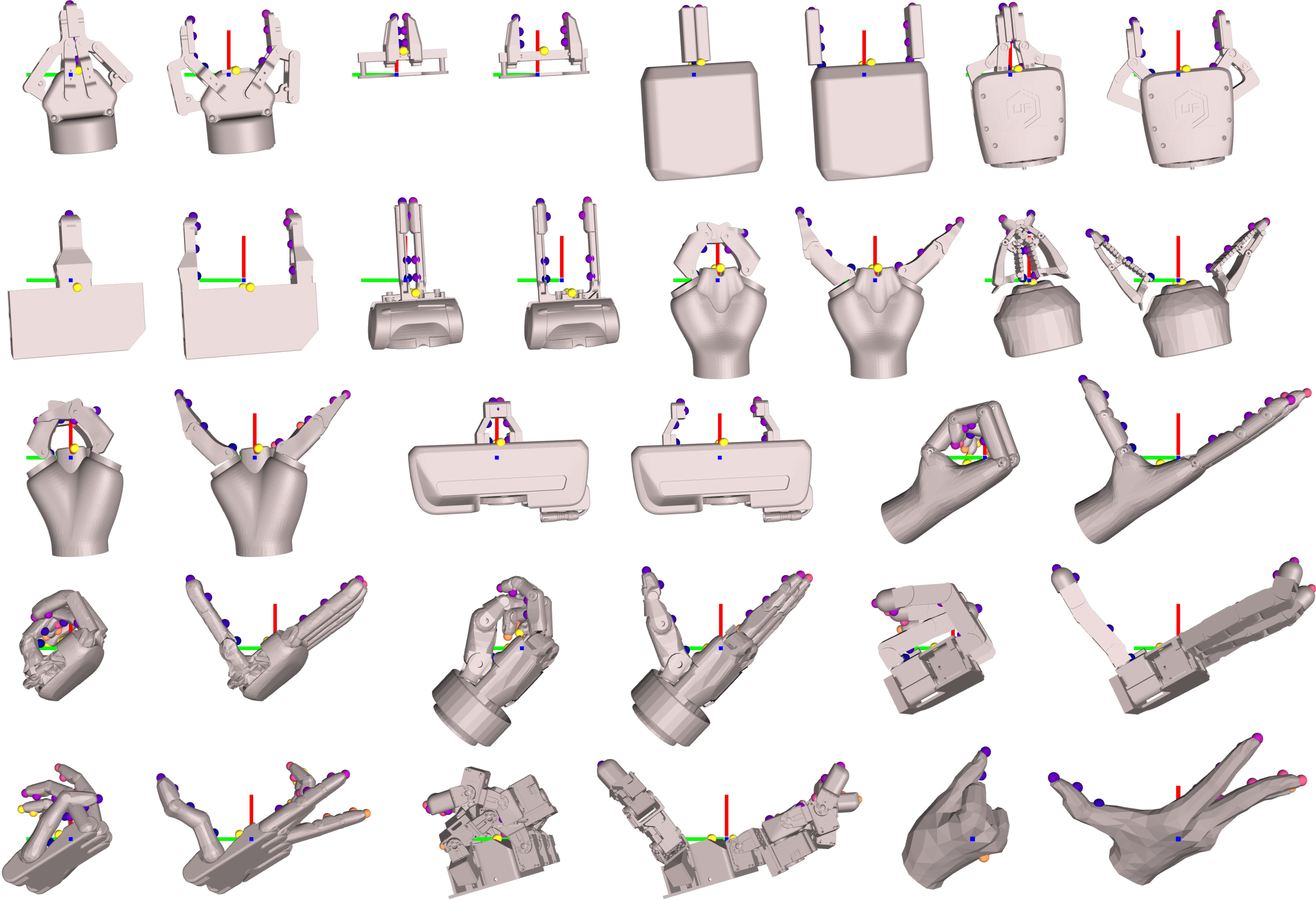}
  \end{center}
  \vspace{-1mm}
  \caption{Manipulators are configured using only the first principal component $1^{\text{st}}$pc = \{-3, 3\}, in their own refined end-effector frame: \textit{Robotiq}, \textit{WidowX}, \textit{Fetch}, \textit{xArm}, \textit{WSG50}, \textit{Rethink}, \textit{Kinova2F}, \textit{GoogleBot}, \textit{Kinova3F}, \textit{Franka}, \textit{Armar}, \textit{ergoCub}, \textit{Schunk}, \textit{Allegro}, \textit{Shadow}, \textit{LEAP}, and \textit{MANO}. }
  \label{fig:pose_pc}
  \vspace{-3mm}
\end{figure}

We qualitatively examine the latent space and the effect that the \acrshort{pca} coefficients have on manipulator poses. 
~\cref{fig:pose_pc} shows the poses of manipulators at the two ends of the first principal component. Setting $1^{\text{st}}$pc = 3 corresponds to fully open and $1^{\text{st}}$pc = -3 corresponds to fully closed configurations, consistently across all 16 rigid manipulators and one non-rigid \textit{mano} hand. Moreover, each end-effector frame is shifted from the palm center and oriented differently but coherently across manipulators. This shows that PCHands provides a unified representation for a variety of manipulators in both pose synergies and end-effector frames.

\subsection{Dexterous Manipulation with \acrshort{rl}}
\label{sec:benchmark}

\begin{figure*}[t]
  \vspace{2mm}
  \begin{center}
    \includegraphics[width=1\linewidth]{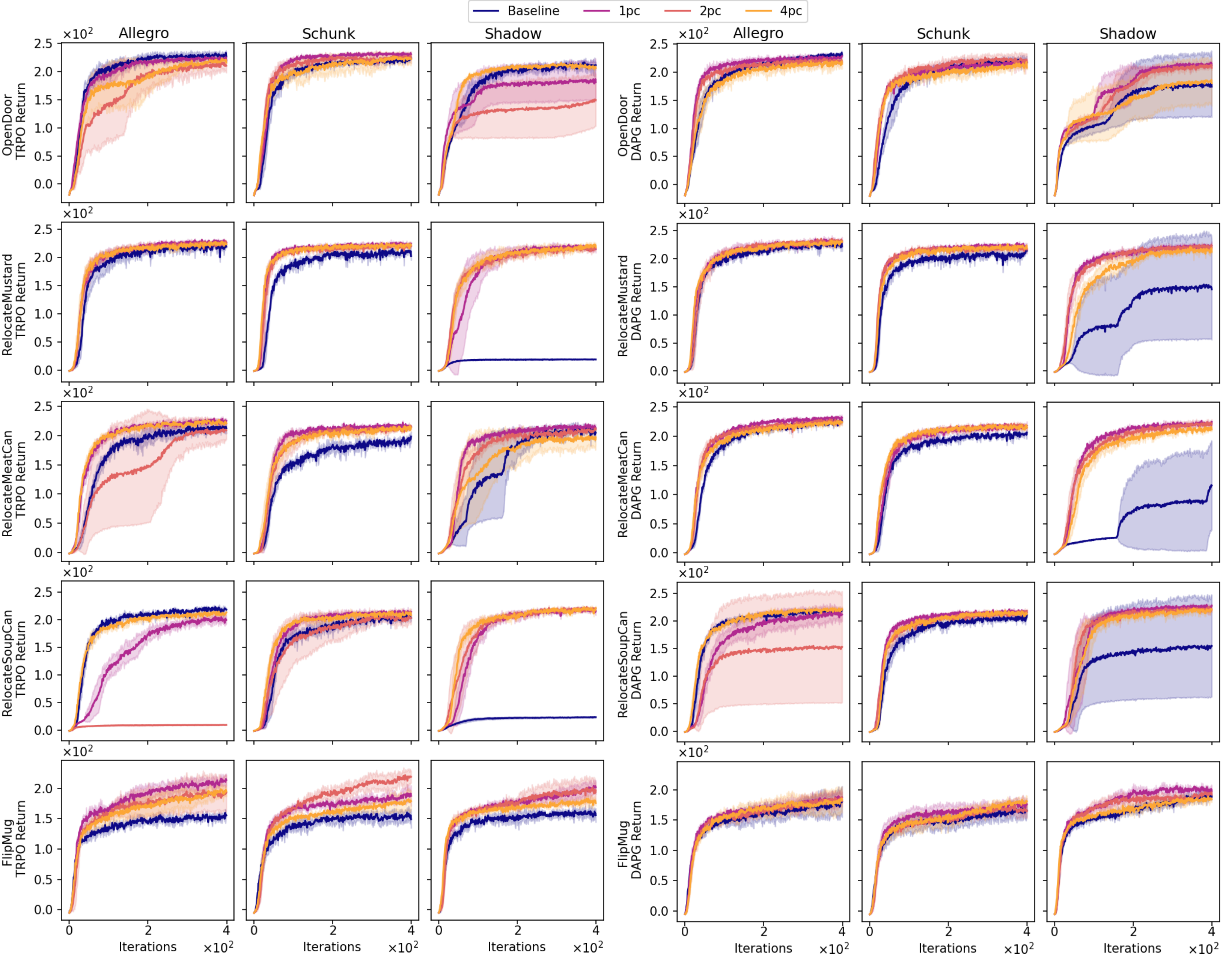}
  \end{center}
  \vspace{-2mm}
  \caption{Average episode return over training of 2 \acrshort{rl} algorithms, 3 manipulators and 5 tasks. We repeat all experiments with 3 different seeds and report the mean and standard deviation to compare PCHands (\textit{N}-pc) to our baseline~\cite{qin2022one}.}
  \label{fig:rl_baseline}
  \vspace{-3mm}
\end{figure*}

We evaluate PCHands on solving dexterous manipulation tasks in \acrshort{rl} settings. Similar to~\cite{qin2022one}, we benchmark results on 5 tasks (\textit{Open-Door}, \textit{Relocate-Mustard}, \textit{Relocate-MeatCan}, \textit{Relocate-SoupCan}, and \textit{Flip-Mug}) and consider 3 anthropomorphic manipulators (\textit{Allegro}, \textit{Schunk}, and \textit{Shadow} hand). 
We use~\cite{qin2022one} as our baseline method, where the policy observes and controls the target manipulator in the joint space. 
In contrast, PCHands policy uses the first \textit{N} principal components (\textit{N}-pc) of $z'$ in both the observation and action spaces, as described in Sec.~\ref{sec:synergy}. Both the baseline and PCHands adopt the \textit{flying-hand} mode: besides controlling fingers' configuration, they also control the end-effector pose $T^\text{eef}$ in Cartesian space. Differing from the original implementation in~\cite{qin2022one}, we replace velocity control with position control mode for better learning stability.
We benchmark the tasks using two \acrshort{rl} algorithms: Trust Region Policy Optimization (TRPO)~\cite{Schulman2015trpo} and Demo Augmented Policy Gradient (DAPG)~\cite{Rajeswaran2018dapg}. 
Both TRPO and DAPG are on-policy algorithms. While TRPO optimizes the policy without any prior information, DAPG utilizes task demonstrations. 

\noindent{\textbf{Human Demonstrations}}
We gather task demonstrations for DAPG via teleoperation in Sapien~\cite{Xiang_2020_SAPIEN} by retargeting human hand poses (in MANO hand model~\cite{romero2017embodied}) to the \textit{customized-hand} presented in~\cite{qin2022one}, which is a robotic hand customized to closely fit the shape and size of the human demonstrator. In doing this, we use PCHands for retargeting as explained in Sec.~\ref{sec:synergy}.
This allows us to gather 50 successful demonstrations per task with a source manipulator $\mathrm{src}$ (\textit{customized-hand}). 
For the baseline, hand configurations and end-effector pose in the demonstrations are retargeted into each target manipulator $\mathrm{tgt}$ (respectively obtaining $T^\text{eef}_\text{tgt}$ and $j_\mathrm{tgt}$) by using \acrshort{adf} anchors, and directly finding joint values by optimizing~\cref{equ:optim} as in~\cite{qin2022one, qin2023anyteleop}. 
For PCHands instead, we extract the \textit{N}-pc latent representation from the hand configurations in the demonstrations $z'_\text{tgt} \leftarrow \mathcal{E}(\mathrm{src}, j_\mathrm{src})$ and set the end-effector pose for each target manipulator directly $T^\text{eef}_\text{tgt} \leftarrow T^\text{eef}_\text{src}$.

\noindent{\textbf{Learning Curves}} We show in \cref{fig:rl_baseline} the learning curves for the baseline and PCHands with \textit{N}-pc where $N=\{1, 2, 4\}$ for all 5 tasks and 3 manipulators. 
PCHands converges faster than the baseline in most cases, with (DAPG) and without (TRPO) human demonstrations. This shows that PCHands improves the \acrshort{rl} efficiency by using a \textit{N}-pc latent representation in the observation and action spaces. 
In general, human demonstrations help avoid local optima for both the baseline and PCHands, and for all the manipulators. This result supports the claim that the conversion of the task demonstrations to the latent space of PCHands preserves the useful information. Besides, DAPG performance for \textit{1}-pc and \textit{2}-pc are comparable and predominantly better than \textit{4}-pc, suggesting that 
two \acrshort{dof} may be sufficient to learn the tasks on complex manipulators with high \acrshort{dof} (i.e., 16, 9, 18 respectively for the manipulators in this experiment).

\begin{figure*}[ht]
  \vspace{2mm}
  \centering
  \includegraphics[width=1\textwidth]{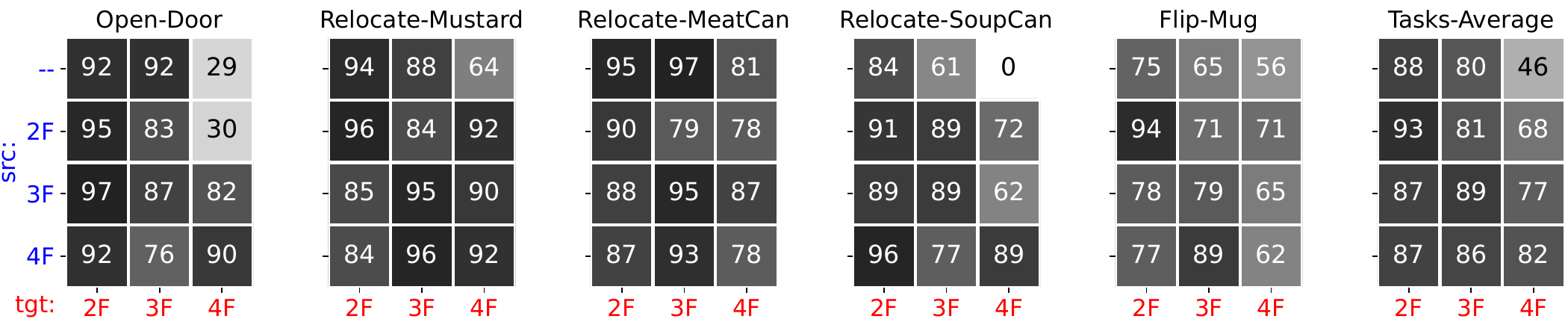}
  \vspace{-5mm}
  \caption{Success rate (\%) of five tasks over target manipulators (\textcolor{red}{2/3/4F}) 
  trained with DAPG using different demonstration sources (\textcolor{blue}{2/3/4F}) and TRPO (\textcolor{blue}{--}). The rightmost table reports the average performance over tasks.}
  \label{fig:rl_demo}
  \vspace{-1mm}
\end{figure*}

\begin{figure}[ht]
  \centering
  \includegraphics[width=1\linewidth]{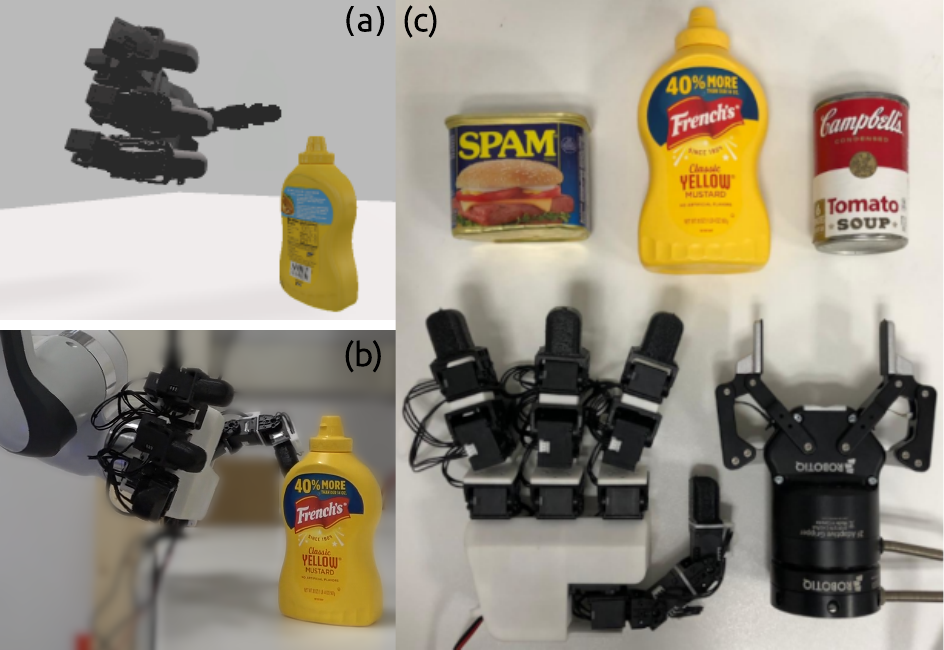}
  \caption{Zero-shot sim-to-real policy transfer: from (a) \textit{flying-hand} mode in simulation to (b) \textit{Cartesian-control} mode on a 7-\acrshort{dof} arm in the real-world, on (c) three YCB~\cite{calli2015ycb} objects and two manipulators.}
  \label{fig:real_exp}
  \vspace{-3mm}
\end{figure}

\subsection{Ablation on the Source of Task Demonstrations}
\label{exp:demo_source}

In this section, we evaluate the performance of PCHands when task demonstrations are collected with a different manipulator. 
To this end, we examine the performance of DAPG on the same five tasks as in Sec.~\ref{sec:benchmark} but using demonstrations from various manipulators. We consider three manipulators: the 2-finger (\textit{2F}) \textit{Robotiq-2f85}, the 3-finger (\textit{3F}) \textit{kinova-3f}, and the 4-finger (\textit{4F}) \textit{LEAP-hand}. 
Instead of collecting demonstrations with the \textit{customized hand} as in Sec.~\ref{sec:benchmark}, for each task we collect 50 demonstrations by teleoperating each manipulator in the Sapien simulator. 
We then extract the \textit{N}-pc latent representation and end-effector trajectory by using PCHands as in Sec.~\ref{sec:benchmark}. We finally train DAPG on the demonstrations from each of the three sources and benchmark performance on each manipulator. Additionally, for each task and manipulator, we train TRPO policies to compare performance against a baseline that does not use task demonstrations for policy learning.

We report in~\cref{fig:rl_demo} the average success rates over 50 trials after 100, 200, 300, and 400 training steps for all combinations of demonstration sources. In this experiment, we consider \textit{2}-pc based on the findings from Sec.~\ref{sec:benchmark}.
Apart from observing the expected result that high performance is achieved when the demonstrations are collected on the same target manipulator, we observe that DAPG achieves consistently higher performance than TRPO even when using demonstrations from a different source, especially when the target is a dexterous manipulator (see the \textit{4F-tgt} in the task average performance). This supports the claim that the proposed PCHands is a viable and efficient approach to re-use available demonstrations, even if for different manipulators.

\subsection{Transfer to Real-World Setup}

In this section, we examine the roll-out of the DAPG policies learned with PCHands in simulation in Sec.~\ref{exp:demo_source} on a real robot. 
As depicted in \cref{fig:real_exp}, the considered platform includes a 7-\acrshort{dof} \textit{Franka-Panda} robotic arm equipped with either a \textit{Robotiq-2f85} (\textit{2F} in~\cref{tab:real}) or a \textit{LEAP-hand} (\textit{4F}), and an external RGB-D camera (a \textit{RealSense L515}). To fulfill the observation space of the policy, $T^{eef}$, $N$-pc and the object 6D pose $T^{obj}$ are needed. We get $T^{eef}$ and the $N$-pc manipulator configuration with proprioception, and use FoundationPose~\cite{wen2024foundationpose} as a 6D object tracker that provides the pose of the target object from RGB-D images. We then use the \textit{libfranka} motion generator to map the output of our policy in the form of 6D end-effector poses to joint trajectories. 
\cref{tab:real} reports the success rate over 10 roll-outs achieved by the policies trained for 400 steps in Sec.~\ref{exp:demo_source} on the three \textit{Relocate} tasks, deployed zero-shot on the real robot without any sim-to-real adaptation.

The real-world performance remains comparable to simulation results in the \textit{Relocate-Mustard} task for both \textit{2F} and \textit{4}F manipulators, and in the \textit{Relocate-MeatCan} and \textit{Relocate-SoupCan} tasks for the \textit{2F} manipulator. 
The performance degradation observed with the \textit{4F} manipulators in the latter tasks can be attributed to the significant occlusion on the object caused by \textit{4-finger} encompassing grasp, which hinders the vision-based object pose tracker and yields inaccurate object pose estimates.
This is particularly evident in the \textit{Relocate-SoupCan} experiment using the \textit{LEAP-Hand}. 
In contrast, the ground-truth object pose is always available in simulation, which explains the performance gap between simulated and real-world experiments.

\begin{table}
    \caption{Success rate (\%) of real-world roll-outs.}
    \vspace{-1mm}
    \centering
    \begin{tabular}{c|c|ccc|c}
        \toprule
        \multirow{2}{*}{Task} & Target & \multicolumn{3}{c|}{Demo Source} & \multirow{2}{*}{Average}\\ 
        & Manip. & 2F & 3F & 4F & \\
        \midrule
        \multirow{2}{*}{Relocate-Mustard} & 2F & 90 & 100 & 100 & 97 \\
        & 4F & 100 & 80 & 90 & 90 \\
        \midrule
        \multirow{2}{*}{Relocate-MeatCan} & 2F & 100 & 80 & 90 & 90 \\
        & 4F & 50 & 30 & 70 & 50 \\
        \midrule
        \multirow{2}{*}{Relocate-SoupCan} & 2F & 80 & 80 & 70 & 77 \\
        & 4F & 70 & 50 & 0 & 40 \\
        \bottomrule
    \end{tabular}
    \label{tab:real}
    \vspace{-3mm}
\end{table}

\section{Conclusion}
Extracting a unified representation across human and robotic manipulators is essential for effective data and policy transfer, enabling scalable training of generalist robotic models. To this end, we introduce PCHands, a framework that derives hand postural synergies for the human hand and 16 diverse robotic manipulators. PCHands utilizes the proposed \acrshort{adf} to: (i) extract latent manipulator representations via a \acrshort{cvae}, (ii) reduce task and joint dimensionality using \acrshort{pca}, and (iii) align end-effector frames across manipulators with \acrshort{icp}. 
Compared to optimization-based retargeting baselines that learn tasks in joint space, PCHands demonstrates greater efficiency in \acrshort{rl}-based manipulation. Furthermore, PCHands supports robust learning from demonstrations across different sources. 
Real-world experiments confirm that PCHands policies trained in simulation transfer directly to physical manipulators in relocation tasks without requiring sim-to-real adaptation. 
Future work aims to expand the coverage to a larger, or even open, set of manipulators and leverage large-scale public datasets collected via two-finger grippers or human demonstrations to train policies for more complex dexterous robotic tasks.

\section*{Acknowledgements}
This research was supported by the PNRR MUR project PE0000013-FAIR, by the National Institute for Insurance against Accidents at Work (INAIL) project ergoCub-2.0, and by the Brain and Machines Flagship Programme of the Italian Institute of Technology.

\bibliography{bib}

\end{document}